\documentclass[10pt,twocolumn,letterpaper]{article}
\pdfoutput=1
\usepackage{iccv}
\usepackage{times}
\usepackage{epsfig}
\usepackage{graphicx}
\usepackage{amsmath}
\usepackage{amssymb}
\usepackage{dsfont}

\usepackage[pagebackref=true,breaklinks=true,letterpaper=true,colorlinks,bookmarks=false]{hyperref}

\iccvfinalcopy 

\def\iccvPaperID{2006} 

\ificcvfinal\fi
\begin{document}

\title{PPR-FCN: Weakly Supervised Visual Relation Detection via Parallel Pairwise R-FCN}

\author{Hanwang Zhang$^{\dagger}$, Zawlin Kyaw$^\ddagger$, Jinyang Yu$^\dagger$, Shih-Fu Chang$^\dagger$\\
	$^\dagger$Columbia University, $^\ddagger$National University of Singapore\\
	\{hanwangzhang, kzl.zawlin, yjy941124\}@gmail.com; {shih.fu.chang@columbia.edu}\\
}

\maketitle

\begin{abstract}
We aim to tackle a novel vision task called Weakly Supervised Visual Relation Detection (WSVRD) to detect ``subject-predicate-object'' relations in an image with object relation groundtruths available only at the image level. This is motivated by the fact that it is extremely expensive to label the combinatorial relations between objects at the instance level. Compared to the extensively studied problem, Weakly Supervised Object Detection (WSOD), WSVRD is more challenging as it needs to examine a large set of regions pairs, which is computationally prohibitive and more likely stuck in a local optimal solution such as those involving wrong spatial context. To this end, we present a Parallel, Pairwise Region-based, Fully Convolutional Network (PPR-FCN) for WSVRD. It uses a parallel FCN architecture that simultaneously performs pair selection and classification of single regions and region pairs for object and relation detection, while sharing almost all computation shared over the entire image. In particular, we propose a novel position-role-sensitive score map with pairwise RoI pooling to efficiently capture the crucial context associated with a pair of objects. We demonstrate the superiority of PPR-FCN over all baselines in solving the WSVRD challenge by using results of extensive experiments over two visual relation benchmarks.
\end{abstract}

\section{Introduction}
Visual relation detection (VRD) aims to detect objects and predict their relationships in an image, especially \texttt{subject}-\texttt{predicate}-\texttt{object} triplets like \texttt{person}-\texttt{hold}-\texttt{ball} (verb), \texttt{dog}-\texttt{on}-\texttt{sofa} (spatial), \texttt{car}-\texttt{with}-\texttt{wheel} (preposition), and \texttt{person1}-\texttt{taller}-\texttt{person2} (comparative)~\cite{krishna2016visual}. As an intermediate task between low-level object detection~\cite{li2016r} and high-level natural language modeling~\cite{vinyals2015show}, VRD has received increasing attention recently, in areas of new  benchmarks~\cite{lu2016visual,krishna2016visual}, algorithms~\cite{li2017vip,zhang2016vtranse,Dai_2017_CVPR,Zhang_2017_CVPR}, and visual reasoning~\cite{johnson2015image,Hu_2017_CVPR,wu2016ask}. VRD is expected to become an important building block for the connection between vision and language. 

Like any other visual detection task, VRD is also data-hungry. However, labeling high-quality relation triplets is much more expensive than objects as it requires the tedious inspection of a combinatorial number of object interactions. On the other hand, collecting image-level relation annotation is relatively easier. For example, there are abundant image-caption data~\cite{krishna2016visual,lin2014microsoft} and Web image-text pairs~\cite{thomee2016yfcc100m}, where \emph{image-level} relation descriptions can be automatically extracted from the text using state-of-the-art text parsers~\cite{schuster2015generating, angeli2015leveraging}. Therefore, to make VRD of practical use at a large scale, it is necessary to study the novel and challenging task: \textit{weakly supervised visual relation detection} (WSVRD), with triplet annotation available only at the \emph{image level}.

\begin{figure}
	\centering
	\includegraphics[width=1\linewidth]{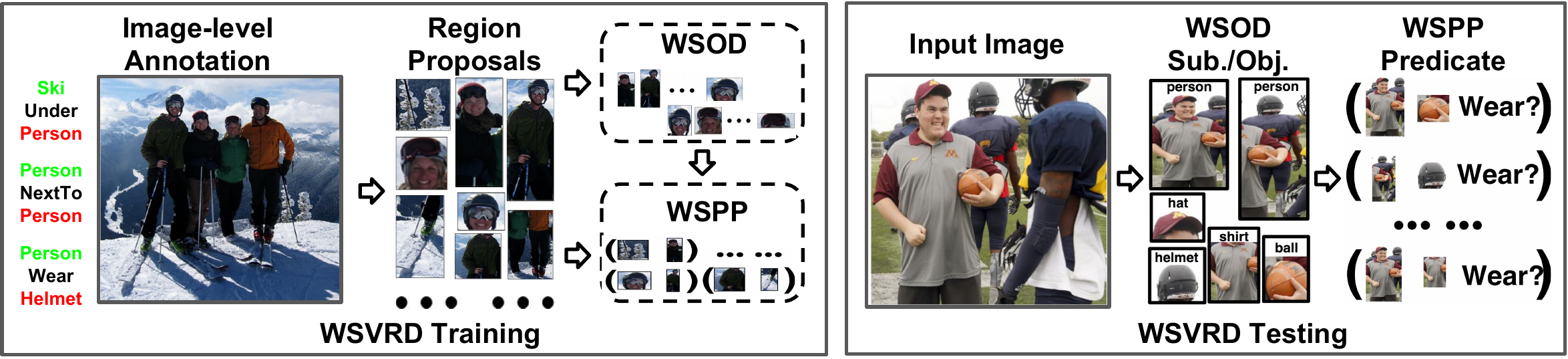}
	\caption{The illustration of WSVRD training and testing stages. It includes weakly supervised object detection (WSOD) and weakly supervised predicate prediction (WSPP). Note that the key difference from WSOD is that WSVRD requires pairwise region modeling for many weakly labeled region pairs in WSPP.}
	\label{fig:1}
\end{figure}

Figure~\ref{fig:1} shows the WSVRD problem studied in this paper. As there are no instance-level object annotations (\eg, bounding boxes), we first exploit region proposal generators~\cite{zitnick2014edge,uijlings2013selective} for a set of candidate proposals (or RoIs) and then predict their object classes. This step is also known as \emph{Weakly Supervised Object Detection} (WSOD)~\cite{bilen2016weakly,cinbis2017weakly}. Then, as the image-level relation does not specify which pairs of objects are related, it exhaustively enumerates every RoI pairs  as candidate \texttt{subject}-\texttt{object} pairs for \texttt{predicate} prediction (\eg, relationships), which results in that WSVRD is more challenging than WSOD. More specifically, first, as the spatial context annotation of pairwise regions is missing, we should carefully model the spatial constraints in WSVRD otherwise the relationships will be easily confused by incorrect \texttt{subject}-\texttt{object} configurations, \eg, one can detect \texttt{hat}-\texttt{on}-\texttt{person} correctly but the \texttt{hat} is on someone else; second, for $N$ regions, WSVRD has to scan through $\mathcal{O}(N^2)$ region pairs, thus, the weakly supervised learning based on alternating between instance selection and classification (\ie, predicate prediction) in WSVRD  is more easily trapped in bad local optimal solution than that in WSOD~\cite{kumar2010self}; third, the $\mathcal{O}(N^2)$ computational cost in WSVRD would become prohibitively expensive if per-RoI fully-connected subnetwork is still adopted~\cite{zhang2016vtranse, Zhang_2017_CVPR}, since WSVRD usually uses many more object regions (\eg, $>$100 regions and $>$10,000 pairs)  than supervised VRD (\eg, $<$20 regions and $<$400 pairs) to ensure high recall of object instances.

\begin{figure*}
	\centering
	\includegraphics[width=.8\linewidth]{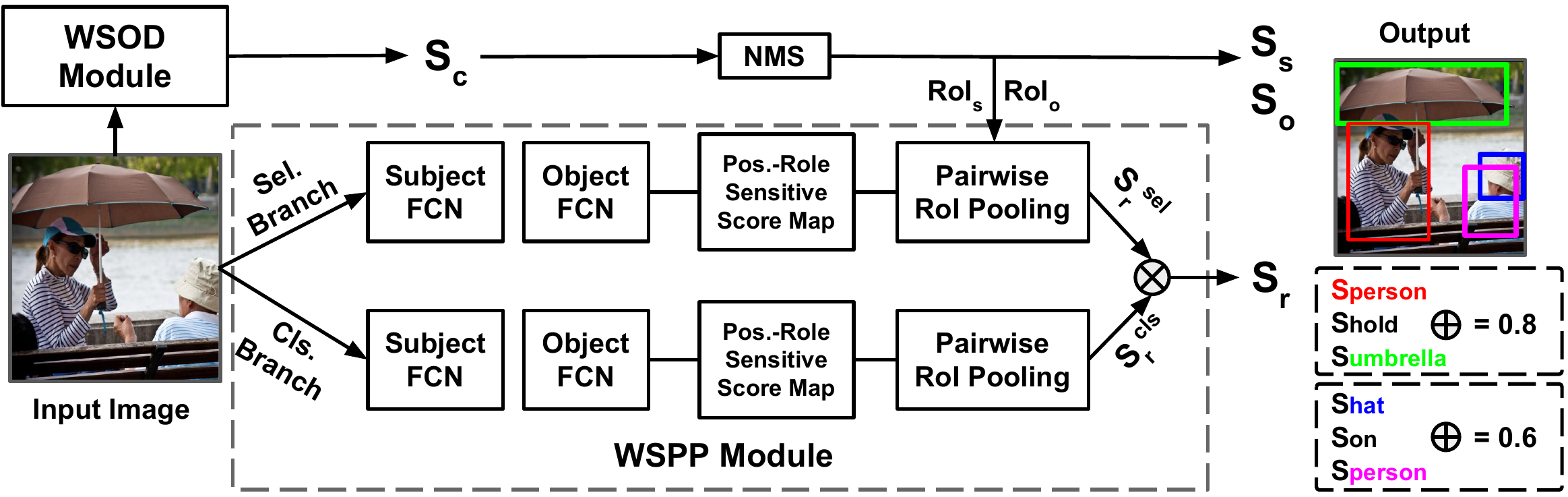}
	\caption{The overview of the proposed PPR-FCN architecture for WSVRD. It has two modules: WSOD for object detection (Section~\ref{sec:3.1}) and WSPP for predicate prediction (Section~\ref{sec:3.2}), each module is composed by a pair selection branch and a classification branch.}
	\vspace{-4mm}
	\label{fig:2}
\end{figure*}

We present a Parallel, Pairwise Region-based, end-to-end Fully Convolutional Network (PPR-FCN) to tackle the above challenges in WSVRD. The architecture is illustrated in Figure~\ref{fig:2} and detailed in Section~\ref{sec:3}. It consists of a WSOD module for weakly supervised object detection and a Weakly Supervised Predicate Prediction (WSPP) module for weakly supervised region pair modeling. PPR-FCN is a two-branch parallel network, inspired by the recent success of using parallel networks to avoid bad local 
optima in WSOD~\cite{bilen2016weakly}. We use FCN~\cite{li2016r} as our backbone network to exploit its advantages in sharing computation over the entire image, making efficient pairwise score estimation possible~\cite{li2016r,long2015fully}. The WSPP module has two novel designs:
\\
\textbf{1. Position-Sequence-Sensitive Score Map}. Inspired by the position-sensitive score map in R-FCN~\cite{li2016r}, we develop a set of \emph{position-role-sensitive} conv-filters to generate a score map, where every spatial pixel encodes the object class-agnostic spatial context (\eg, \texttt{subject} is above \texttt{object} for predicate \texttt{sit on}) and roles (\eg, the first part of the pixel channels is \texttt{subject} and the rest is \texttt{object}). 
\\
\textbf{2. Pairwise RoI Pooling}. To shepherd the training of the position-role-sensitive conv-filters, we append a \textit{pairwise RoI pooling} layer on top of the score map for fast score estimation. Our pooling design preserves the spatial context and subject/object roles for relations. 

To the best of our knowledge, PPR-FCN is the first detection network for the WSVRD task. We believe that PPR-FCN will serve as a critical foundation in this novel and challenging vision task. 


\section{Related Work}
\textbf{Fully Convolutional Networks}. A recent trend in deep networks is to use convolutions instead of fully-connected (fc) layers such as ResNets~\cite{he2015deep} and GoogLeNet~\cite{szegedy2015going}. Different from fc layers where the input and output are fixed size, FCN can output dense predictions from arbitrary-sized inputs. Therefore, FCN is widely used in segmentation~\cite{long2015fully,Liu_2017_CVPR}, image restoration~\cite{eigen2013restoring}, and dense object detection windows~\cite{ren2015faster}. In particular, our PPR-FCN is inspired by another benefit of FCN utilized in R-FCN~\cite{li2016r}: per-RoI computation can be shared by convolutions. This is appealing because the expensive computation of pairwise RoIs is replaced by almost cost-free pooling.

\textbf{Weakly Supervised Object Detection}. As there are no instance-level bounding boxes for training, the key challenge of WSOD is to localize and classify candidate RoIs simultaneously~\cite{cinbis2017weakly,wang2014weakly,song2014learning,Jie_2017_CVPR}. The parallel architecture in PPR-FCN is inspired by the two-branch network of Bilen and Vedaldi~\cite{bilen2016weakly}, where the final detection score is a product of the scores from the parallel localization and classification branches. Similar structures can be also found in \etal~\cite{kantorov2016contextlocnet,rohrbach2016grounding}. Such parallel design is different from MIL~\cite{maron1998framework} in a fundamental way as regions are selected by a localization branch, which is independent of the classification branch. In this manner, it helps to avoid one of the pitfalls of MIL, namely the tendency of the method to get stuck in local optima.

\textbf{Visual Relation Detection}
Modeling the interactions between objects such as verbs~\cite{gupta2008beyond,chao2015hico}, actions~\cite{gupta2009observing,ramanathan2015learning,yao2010modeling}, and visual phrases~\cite{yatskar2016situation,atzmon2016learning,sadeghi2011recognition,desai2012detecting} are not new in literature. However, we are particularly interested in the VRD that simultaneously detects generic \texttt{subject}-\texttt{predicate}-\texttt{object} triplets in an image, which is an active research topic~\cite{lu2016visual,li2017vip,zhang2016vtranse, Zhang_2017_CVPR,Dai_2017_CVPR,lisce2017} and serves as a building block for connecting vision and language~\cite{krishna2016visual,johnson2015image,Hu_2017_CVPR,wu2016ask}. But, a key limitation is that it is very expensive to label relation triplets as the complexity is combinatorial. Perhaps the most related work to ours is done by Prest \etal~\cite{prest2012weakly} on weakly-supervised learning human and object interactions. However, their spatial configurations and definitions of relations are limited to one person and one object while our relations include  generic objects and diverse predicates. There are recent works on referring expression groundings, \eg, localizing an object by its relationship to another object~\cite{hu2016modeling,yu2016modeling,nagaraja2016modeling}. However, they require stronger supervision, \ie, at least one of the objects is labeled with bounding box. We also notice that we are not the only work towards the efficiency of VRD. Li \etal~\cite{li2017vip} and Zhang \etal~\cite{Zhang_2017_CVPR} proposed to use groundtruth pairwise bounding boxes to learn triplet proposals to reduce the number of region pairs; however, these methods are fully supervised. 

\section{PPR-FCN}\label{sec:3}
As illustrated in Figure~\ref{fig:2}, PPR-FCN consists of two modules: 1) WSOD module for object detection and 2) WSPP module for predicate prediction. At test time, PPR-FCN first detects a set of objects and then predicts the predicate for every object pairs. In this section, we will detail each module.

\subsection{WSOD Module}\label{sec:3.1}
The goal of WSOD module is to predict the object class score $S_c(P)$ for a RoI $P$ of any class $c\in\{0,1,...,C\}$. Then, NMS is performed to generate the final detection result: subject and object RoIs (\ie, bounding boxes), and their classes. It is worth noting that any state-of-the-art WSOD method can be used as the WSOD module in PPR-FCN. In this paper, we adopt a parallel design similar in WSDDN~\cite{bilen2016weakly} as it achieves the state-of-the-art results on benchmarks (cf. Section~\ref{sec:4.3}) and it is easy to replace the backbone network with R-FCN~\cite{li2016r}, which is also compatible with the subsequent WSPP module. During training and test, we first use EdgeBox~\cite{zitnick2014edge} to generate $N$ object RoIs, where $N$ is initially 4,000 and then reduced to 1,000 by NMS with IoU$>$0.4 and discard those with objectness score$<$0.2; the objectness score for a region is the sum over all class scores from a 3-epoch pre-trained WSOD with the initial 4,000 RoIs. Then, given the 1,000 RoIs, for each class, we perform NMS with IoU$>$0.4 and score threshold$>$0.7 to select 15$\sim$30 RoIs, resulting $\sim$100 detected objects, where this number is significantly larger than that in supervised detection (\eg, $\sim$20), since we need to ensure enough recall for true objects. Please see Section~\ref{sec:3.3} for the training loss of this module.

\subsection{WSPP Module}\label{sec:3.2}
WSPP module predicts the predicate score $S_r(P_i,P_j)$ of any predicate $r\in\{1, ..., R\}$ for two RoIs detected by the previous WSOD module. As shown in Figure~\ref{fig:2}, it is a two-branch network with independent parameters for pair selection (\ie, which pair of regions are related) and classification. In particular, the input feature map for WSPP is the same as WSOD, which is the base CNN feature map followed by a trainable conv-layer as in R-FCN~\cite{li2016r}. The predicate score, \ie, the likelihood of \texttt{subject}-\texttt{object} pair being associated with predicate $r$, is defined as:
\begin{equation}\label{eq:1}
S_r(P_i,P_j) = S^{sel}_r(P_i,P_j)\cdot S^{cls}_r(P_i,P_j),
\end{equation}
where we split the challenging estimation of the predicate score using only image-level annotation into two simpler problems: one is responsible for pair selection and the other is for predicate classification. In particular, $S^{sel}_r$ (or $S^{cls}_r$) is the predicate score from the selection (or classification) branch. $S^{sel}_r$ is softmax normalized over all possible region pairs with respect to a predicate class, \ie, $S^{sel}_r(P_i, P_j)\leftarrow \textrm{softmax}_{i,j}S^{sel}_r(P_i, P_j)$; while $S^{cls}_r$ is softmax normalized over possible predicate classes for a region pair, \ie, $S^{cls}_r(P_i, P_j) \leftarrow \textrm{softmax}_{r}S^{cls}_{r}(P_i, P_j)$. Note that such normalizations assign different objectives to two branches and hence they are unlikely to learn redundant models~\cite{bilen2016weakly}. Essentially, the normalized selection score can be considered as a soft-attention mechanism used in weakly supervised vision tasks~\cite{rohrbach2016grounding, Chen_2017_CVPR} to determine the likely RoIs. Next, we will introduce how to calculate the scores before normalization. Without loss of generality, we use $S^{cls}_r$ as the example and discard the superscript. 

\subsubsection{Position-Sequence-Sensitive Score Map}
First, predicate score should be position-sensitive as the spatial context of two objects is informative for the relationship. Second, as the predicate score is usually dependent on the role-sequence of two RoIs, the score should be also role-sensitive to ensure asymmetric scores of $S_r(P_i, P_j)$ and $S_r(P_j, P_i)$. For example, for \texttt{ride} score, \texttt{person}-\texttt{ride}-\texttt{bike} is more likely than \texttt{bike}-\texttt{ride}-\texttt{person};  \texttt{person}-\texttt{on}-\texttt{bike} is different from  \texttt{bike}-\texttt{on}-\texttt{person}, as the former usually indicates ``person riding a bike'' while the latter suggests ``person carrying a bike''. Inspired by the \emph{position-sensitive} score map design in R-FCN~\cite{li2016r}, we propose to use two sets of trainable size $1\times 1$ and stride 1 conv-filters to generate $2\cdot k^2R$-channel \emph{position-role-sensitive} score maps from the input feature map. As illustrated in Figure~\ref{fig:3}, the first $k^2R$-channel score map encodes $R$ predicate scores at $k^2$ spatial positions for \texttt{subject} and the second $k^2R$-channel map encodes scores for \texttt{object}. By using these filters, the computation of predciate prediction is amortized over the entire image. Note that the score maps are class-agnostic, \ie, they are only aware of whether a spatial location is \texttt{subject} or \texttt{object} but not aware of whether it is ``dog'' or ``car''. This is scalable to relation detection for many classes and predicates as the complexity is only $\mathcal{O}(C+R)$ but not $\mathcal{O}(C^2R)$.

\subsubsection{Pairwise RoI Pooling}
\begin{figure}
	\centering
	\includegraphics[width=1\linewidth]{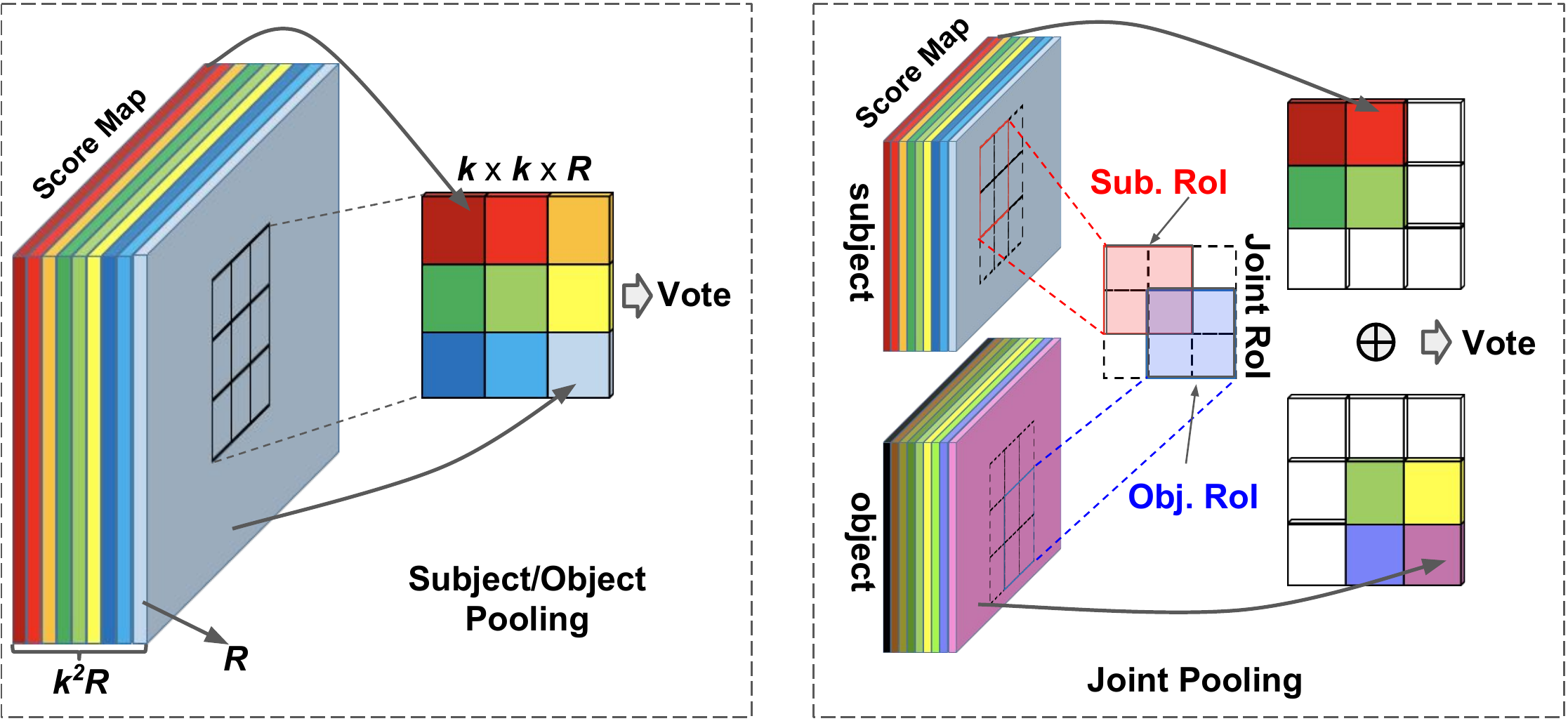}
	\caption{Illustrations of pairwise RoI pooling with $k^2 (k = 3)$ spatial grids and $R$ predicates. \textbf{Left}: subject/object pooling for a single RoI. \textbf{Right}: joint pooling. For each score map, we use $k^2$ colors to represent different position channels. Each color channel has $R$ predicate channels.  For the joint pooling, uncolored pooling results indicate zero and the back-propagation is disabled through these grids. Note that the score maps in subject/object pooling and joint pooling are different, \ie, there are $2\cdot 2\cdot k^2R$ conv-filters.}
	\label{fig:3}
	\vspace{-4mm}
\end{figure}

To sheperd the training of the above position-role-sensitive filters, we design a \emph{pairwise RoI pooling} strategy to obtain the predicate score $S_r(P_i, P_j)$ for a RoI pair. It includes three pooling steps: 1) subject pooling, 2) object pooling, and 3) joint pooling. Thus, the final $S_r(P_i, P_j)$ is the sum of these steps:
\begin{equation}\label{eq:2}
S_r(P_i,P_j) = S^{sub}_r(P_i) + S^{obj}_r(P_j) + S^{joint}_r(P_i, P_j).
\end{equation}
Next, we will detail the three pooling steps as illustrated in Figure~\ref{fig:3}. 
\\
\textbf{Subject/Object Pooling}. This pooling aims to score whether an RoI is \texttt{subject} or \texttt{object} in a relation. Without loss of generality, we use subject pooling as the walk-through example. We first divide the RoI $P$ into $k\times k$ spatial grids. Suppose $(x,y)\in g(i,j)$ is the set of pixels within the grid $g(i,j)\in P$, where $1\le i, j \le k$, and $X_{x,y,g(i,j),r}$ is the score of the $r$-th predicate at the position $(x,y)$ inside grid $g(i,j)$ in the subject score map $X$, then the subject pooling for $P$ is defined as:
\begin{equation}\label{eq:3}
S^{sub}_r(P)=\underset{g(i,j)\in P}{\operatorname{vote}}\left(\underset{(x,y)\in g(i,j)}{\operatorname{pool}}\left(X_{x,y,g(i,j),c}\right)\right),
\end{equation}
where $k = 3$, $\textrm{pool}(\cdot)$ is mean pooling, and $\textrm{vote}(\cdot)$ is average voting (\eg, average pooling for the scores of the grids). $S^{sub}_r(P)$ is position-sensitive because Eq.~\eqref{eq:3} aggregates responses for a spatial grid of RoI subject to the corresponding one from the $k^2$ maps (\eg, in Figure~\ref{fig:3} left, the dark red value pooled from the top-left grid of the RoI) and then votes for all the spatial grids. Therefore, the training will shepherd the $k^2R$ subject filters to capture subject position in an image.  
\\
\textbf{Joint Pooling}. The above subject/object pooling does not capture the relative spatial context of a predicate. Therefore, we use joint pooling to capture how two RoIs interacts with respect to a predicate. As shown in Figure~\ref{fig:3} right, different from the single RoI pooling where the $k^2$ spatial grids are over the entire RoI, the pairwise RoI pooling is based on the grids over the joint region and the pooling result for the \texttt{subject} $P_i$ or \texttt{object} $P_j$ is from the intersected grids between $P_i$ (or $P_j$) and $P_i\cup P_j$, where the latter joint RoI is divided into $k\times k$ spatial grids. Denote $(x,y)\in g(i',j')$ as the pixel coordinates within the grid $g(i',j') \in P_i\cup P_j$, where $1\le i',j'\le k$, and $X^s_{x,y,g(i',j'),r}$ (or $X^o_{x,y,g(i',j'),r}$) is the score of the $r$-th predicate at the position $(x,y)$ within $g(i',j')$ from the \texttt{subject} (or \texttt{object}) score map. Therefore, the joint RoI pooling is defined as:
\begin{equation}\label{eq:4}
\begin{array}{l}
S^{joint}_r(P_i,P_j)=\!\!\!\!\!\!\!\!\underset{g(i',j')\in P_i\cup P_j}{\operatorname{vote}}\left(\underset{(x,y)\in g(i',j')\cap P_i}{\operatorname{pool}}\!\!\!\left(X^s_{x,y,g(i',j'),r}\right)\right.\\
+\left.\underset{(x,y)\in g(i',j')\cap P_j}{\operatorname{pool}}\left(X^o_{x,y,g(i',j'),r}\right)\right),
\end{array}
\end{equation}
where $g(i',j')\cap P_i$ denotes the intersected pixels between $g(i',j')$ and $P_i$; in particular, if $g(i',j')\cap P_i = \phi$, $\textrm{pool}(\cdot)$ is zero and the gradient is not back-propagated. We set $k = 3$, $\textrm{pool}(\cdot)$ to average pooling, and $\textrm{vote}(\cdot)$ to average voting. For example, for relation \texttt{person}-\texttt{ride}-\texttt{bike}, the pooling result of \texttt{person} RoI is usually zero at the lower grids of the joint RoIs while that of \texttt{bike} RoI is usually zero at the upper grids.

\subsection{Loss Functions}\label{sec:3.3}
We follow the conventional image-centric training strategy~\cite{ren2015faster}, \ie, a training mini-batch arises from the set of region proposals in a single image. We resized images to the longer side of 720 pixels. Multiple scales at \{0.6,0.8,1.0,1.2,1.4\} and random horizontal flips are applied to the images during training. 
\\
\textbf{WSOD Loss}.
Suppose $\mathcal{C}$ is the set of image-level object class groundtruth, $S_c = \sum_iS_c(P_i)$ is the image-level class score\footnote[2]{Note that $S_c$ is also a sum of element-wise product of softmax normalized scores, \ie, $S_c(P_i) = S^{loc}_c(P_i)\cdot S^{cls}_c(P_i)$ and thus it is $<1$.} , the loss is defined as: 
\begin{equation}\label{eq:5}
\mathcal{L}^{obj}_{img} = -\sum\limits^{C}_{c=1}\left(\mathds{1}_{[c \in \mathcal{C}]}\log S_c  + \mathds{1}_{[c \notin \mathcal{C}]}\log\left(1-S_c\right)\right),
\end{equation}
where $\mathds{1}_{[x]}$ is 1 if $x$ is true and 0 otherwise. However, the above image-level loss does not guarantee the spatial smoothness of detection scores. Inspired by the positive and negative bounding box sampling in supervised object detection~\cite{girshick2015fast}, we regularize the smoothness as: 1) for each foreground class $c\in\{1,...,C\}$, the top high-scored regions (\eg, top 5) and their neighborhood with $\textrm{IoU}\ge 0.5$ should both have high scores; we consider them as the pseudo positive regions; and 2) the neighborhood of the pseudo positive regions with $0.1\le \textrm{IoU} \le 0.5$ should be pseudo background regions ($c=0$). In this way, our spatial smoothness regularization loss is:
\begin{equation}\label{eq:6}
\mathcal{L}^{obj}_{reg} = -\sum\limits_{c\in\mathcal{C}}\sum\limits_{i\in\mathcal{C}_c}\log S_c(P_i)-\sum\limits_{i\in\mathcal{B}}\log S_0(P_i),
\end{equation}
where $\mathcal{C}_c$ is the set of pseudo positive regions for class $c\neq 0$, and $\mathcal{B}$ is the set of pseudo background regions. We follow a similar sampling strategy as in~\cite{ren2015faster}: 256 regions are sampled, where at least 75\% are the pseudo background regions.
\\
\textbf{WSPP Loss}.
Suppose $\mathcal{R}$ is the set of the image-level relation groundtruth triplets, specifically, $(s,r,o)\in \mathcal{R}$, where $s,o\in\{1,...,C\}$ are the labels of the \texttt{subject} and \texttt{object}. Suppose $\mathcal{C}_s$ and $\mathcal{C}_o$ are the region sets of \texttt{subject} $s$ and \texttt{object} $o$, respectively. Denote $S_r = \sum_{i\in\mathcal{C}_s,j\in\mathcal{C}_o}S_r(P_i,P_j)$ as the image-level predicate score, the image-level predicate prediction loss is defined as:
\begin{equation}\label{eq:7}
\mathcal{L}^{pred}_{img}\!\! =\! -\!\!\sum\limits_{r = 1}^{R}\!\!\left(\mathds{1}_{[(s,r,o)\in \mathcal{R}]}\log S_r\! +\!\mathds{1}_{[(s,r,o)\notin \mathcal{R}]}\log (1\!-\!S_r)\right) .
\end{equation}
\\
\textbf{Overall Loss}.
The overall loss of PPR-FCN is a multi-task loss that consists of the above WSOD and WSVRD losses:
\begin{equation}\label{eq:8}
\mathcal{L}_{PPR-FCN} = \mathcal{L}^{obj}_{img} + \mathcal{L}^{pred}_{img} + \alpha\mathcal{L}^{obj}_{reg},
\end{equation}
where $\alpha$ is empirically set to 0.2. We train the PPR-FCN model by SGD with momentum~\cite{kingma2014adam}.

\section{Experiments}
\subsection{Datasets}
We used two recently released datasets with a wide range of relation annotation. Every image from the above two datasets is annotated with a set of \texttt{subject}-\texttt{predicate}-\texttt{object} triplets, where every instance pair of \texttt{subject} and \texttt{object} is labeled with bounding boxes. At training time, we discarded the object bounding boxes to conform with the weakly supervised setting.
\\ 
\textbf{VRD}: the Visual Relationships Dataset collected by Lu \etal~\cite{lu2016visual}. It contains 5,000 images with 100 object classes and 70 predicates, resulting in 37,993 relation annotations with 6,672 unique relations and 24.25 predicates per object class. We followed the official 4,000/1,000 train/test split.  
\\
\textbf{VG}: the latest Visual Genome Version 1.2 relation dataset constructed by Krishna \etal~\cite{krishna2016visual}. VG is annotated by crowd workers and thus the relations labeling are noisy, \eg, free-language and typos. Therefore, we used the pruned version provided by Zhang \etal~\cite{zhang2016vtranse}. As a result, VG contains 99,658 images with 200 object categories and 100 predicates, 1,174,692 relation annotations with 19,237 unique relations and 57 predicates per object category. We followed the same 73,801/25,857 train/test split.

\subsection{Evaluation Protocols and Metrics}
Since the proposed PPR-FCN has two modules: WSOD and WSPP, we first evaluated them separately and then overall. Thus, we have the following protocols and metrics that are used in evaluating one object detection task~\cite{bilen2016weakly,cinbis2017weakly,kantorov2016contextlocnet} and three relation-related tasks~\cite{lu2016visual,zhang2016vtranse}:
\\
\textbf{1) Object Detection}.
We used the WSOD module trained with image-level object annotations to detect objects in VRD and VG. We followed the Pascal VOC conventions that a correct detection is at least 0.5 IoU with the groundtruth.
\\
\textbf{2) Predicate Prediction}.
Given the groundtruth objects with bounding boxes, we predict the predicate class between every pair of regions. This protocol allows us to study how well the proposed position-role-sensitive score map and pairwise RoI pooling perform without the limitations of object detection. 
\\
\textbf{3) Phrase Detection}.
We predict a relation triplet with a bounding box that contains both \texttt{subject} and \texttt{object}. The prediction is correct if the predicted triplet is correct and the predicted bounding box is overlapped with the groundtruth by IoU$>$0.5. 
\\
\textbf{4) Relation Detection}.
We predict a relation triplet with the \texttt{subject} and \texttt{object} bounding boxes. The prediction is correct if the predicted triplet is correct and both of the predicted \texttt{subject} and \texttt{object} bounding boxes are overlapped with the groundtruth by IoU$>$0.5. 

Note that both the objects and relations in VRD and VG are not completely annotated. Therefore, the popular Average Precision is not a proper metric as the incomplete annotation will penalize the detection if we do not have that particular groundtruth\footnote[3]{For example, even though R-FCN is arguably better than than Faster R-CNN,  R-FCN only achieves 6.47\% mAP while Faster-RCNN achieves 13.32\% mAP on VRD.}. To this end, following~\cite{lu2016visual,zhang2016vtranse}, we used Recall@50 (\textbf{R@50}) and Recall@100 (\textbf{R@100}) for evaluation. R@K computes the fraction of times a groundtruth is in the top $K$ confident predictions in an image. 

\subsection{Evaluations of Object Detection}\label{sec:4.3}
\textbf{Comparing Methods}. We compared the proposed WSOD module named \textbf{WSOD} with three state-of-the-art weakly supervised object detection methods: 1) \textbf{WSDDN}~\cite{bilen2016weakly}, the weakly-supervised deep detection network. It has a two-branch localization and classification structure with spatial regularization; 2) \textbf{ContextLocNet}~\cite{kantorov2016contextlocnet}, the context-aware localization network. Besides it is also a two-branch structure, the localization branch is further sub-branched to three context-aware RoI pooling and scoring subnetworks; 3) \textbf{WSL}~\cite{li2016weakly}, the weakly supervised object localization model with domain adaption. It is a two-stage model. First, it filters out the noisy object proposal collection to mine confident candidates as pseudo object instances. Second, it learns a standard Faster-RCNN~\cite{ren2015faster} using the pseudo instances. We used their official source codes as implementations on the VRD and VG datasets in this paper. For fair comparison, we used the same ResNet-50~\cite{he2015deep} as the base network. We also provided the fully-supervised detection model \textbf{R-FCN}~\cite{li2016r} as the object detection upper bound.

\begin{figure*}
	\centering
	\includegraphics[width=1\linewidth]{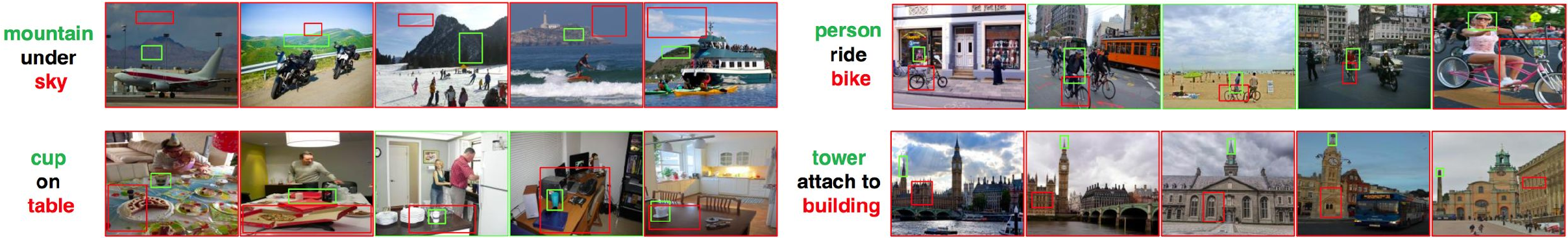}
	\caption{Illustrative top 5 relation detection results from VRD. Red and green borders denote incorrect and correct detections, respectively. Most of the failed cases are due to the wrongly detected objects.}
	\vspace{-2mm}
	\label{fig:5}
\end{figure*}

\textbf{Results}. From Table~\ref{tab:1}, we can see that our WSOD is considerably better than the state-of-the-art methods. This is largely contributed by the parallel FCN architecture. It is worth noting that the quality of the top 1,000 proposal RoIs is significant to WSOD; if we directly used the original scores of EdgeBox, the performance will drop significantly by about 5 points. Note that we are still far behind the fully supervised method such as R-FCN, which shows that there is still a large space to improve WSVRD by boosting WSOD. As illustrated in Figure~\ref{fig:5}, WSOD usually detects the discriminative parts of objects, which is a common failure in state-of-the-art models. We also compared WSOD with other methods on the completely annotated Pascal VOC 2007, where we also achieved the best 39.8\% mAP, surpassing WSDDN (39.3\%), ContextLocNet (36.3\%), and WSL (39.5\%).
\begin{table}[t]
	\centering
	\caption{Weakly supervised object detection performances (R@K\%) of various methods on VRD and VG. The last row is supervised object detection performances by R-FCN.}
	\label{tab:1}
	\scalebox{.8}{
		\begin{tabular}{|c|c|c||c|c|}
			\hline
			Dataset     & \multicolumn{2}{c||}{VRD} & \multicolumn{2}{c|}{VG} \\ \hline
			Metric    & R@50              & R@100            & R@50            & R@100            \\ \hline
			WSDDN~\cite{bilen2016weakly}  & 15.08            & 16.37             & 6.22           & 6.89 \\ \hline
			ContextLocNet~\cite{kantorov2016contextlocnet} & 9.74            & 11.27          & 4.74            & 4.91 \\ \hline
			WSL~\cite{li2016weakly} &13.10             & 13.59          & 5.43            & 6.28 \\ \hline
			WSOD &\textbf{25.34} &\textbf{26.54} &\textbf{8.85} & \textbf{9.12} \\ \hline\hline\hline
			R-FCN~\cite{li2016r} & 57.64            & 58.31          & 18.37         & 18.90 \\ \hline
		\end{tabular}}
	\end{table}

\begin{figure*}
	\centering
	\includegraphics[width=.8\linewidth]{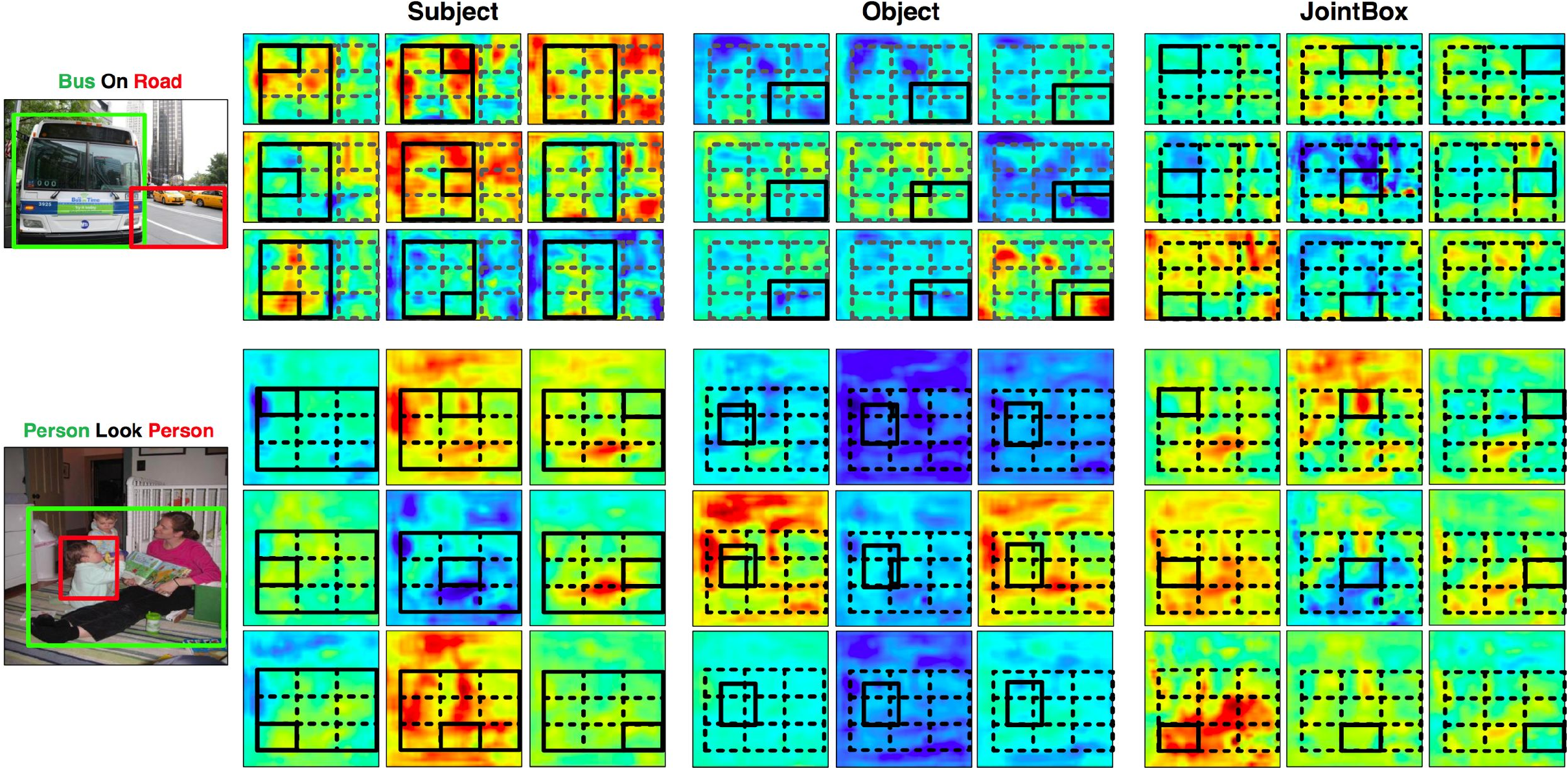}
	\caption{Two illustrative examples (without the base ResNet-50 fine-tuned) of 9 $(k = 3)$ position-role-sensitive score maps trained by pairwise RoI pooling (Subject and Object) and the position-sensitive score maps trained by joint boxes pooling (JointBox). Dashed grids are the joint RoI of \texttt{subject} and \texttt{object}. In Subject and Object score maps, \texttt{subject} and \texttt{object} RoIs are in solid rectangles. The solid grids denotes pooling at corresponding positions. Note that position-role-sensitive pooling is defined as null if the RoI has no overlap with the joint RoI at a position, \eg, \texttt{road} at top left. Please view in color and zoom in.}
	\label{fig:4}
\end{figure*}

\subsection{Evaluations of Predicate Prediction}\label{sec:4.4}
\textbf{Comparing Methods}. Note that the task of predicate prediction is in the supervised setting given the groundtruth of \texttt{subject} and \texttt{object}. Thus, we removed the WSOD module and the localization branch from the WSPP module. In this experiment, our goal is to compare our proposed position-role-sensitive score map and pairwise RoI pooling, namely \textbf{PosSeq+Pairwise} with other three ablated methods: 1) \textbf{Pos-Pairwise} denoting position-sensitive score map followed by pairwise RoI pooling; 2) \textbf{Pos+JointBox} denoting position-sensitive score map followed by joint boxes RoI pooling, where the joint RoI is the tight groundtruth regions that cover both \texttt{subject} and \texttt{object}; and 3) \textbf{PosSeq+Pairwise+fc} denoting position-role-sensitive score map followed by pairwise RoI pooling, but the score is obtained by fully-connected subnetworks. Note that this fc-based method is also comparable to 4) \textbf{VtransE}~\cite{zhang2016vtranse} using the concatenated RoI features from \texttt{subject} and \texttt{object} as the input to its fc prediction network.

\textbf{Results}. From Table~\ref{tab:2}, we can see that our PosSeq+Pairwise outperforms the baselines with non-order score maps and non-pairwise pooling significantly. As illustrated in Figure~\ref{fig:4}, compared to the conventional position-sensitive score maps and pooling, we can observe that PosSeq+Pairwise can capture the contextual configuration better. For example, for the relation \texttt{bus}-\texttt{on}-\texttt{road}, the \texttt{subject} responses are more active at upper positions while the \texttt{object} response are more active at lower positions, and thus the spatial context of \texttt{on} is depicted by adding the pairwise pooling; however, Pos+JointBox seems agnostic to relations but more likely sensitive to objects.

It is worth noting that pooling-based methods are worse than fc-based methods such as VTransE and PosSeq+Pairwise+fc, which contains region-based fully-connected (fc) subnetworks for relation modeling. We noticed that some prepositions such as \texttt{of} and \texttt{by}, and verbs such as \texttt{play} and \texttt{follow}, contain very diverse visual cues and may not be captured by only spatial context. Therefore, fc layers followed by the concatenation of \texttt{subject} and \texttt{object} RoI features might be necessary to model such high-level semantics. Nevertheless, the fact that PosSeq+Pairwise+fc considerably outperforms VTransE demonstrates the effectiveness of exploiting the pairwise spatial context. Note that such unshared region-based subnetworks will lead to inefficient learning in WSPP as there are tens of thousands candidate RoI pairs and millions of fc parameters. 

\begin{table}[t]
	\centering
	\caption{Predicate prediction performances (R@K\%) of various methods on VRD and VG. The last two rows are fc-based methods.}
	\label{tab:2}
	\scalebox{.8}{
		\begin{tabular}{|c|c|c||c|c|}
			\hline
			Dataset     & \multicolumn{2}{c||}{VRD} & \multicolumn{2}{c|}{VG} \\ \hline
			Metric    & R@50              & R@100            & R@50            & R@100            \\ \hline
			Pos+Pairwise  & 24.30            & 24.30             & 43.30           & 43.64 \\ \hline
			Pos+JointBox & 29.57            & 29.57          & 45.69            & 45.78 \\ \hline
			PosSeq+Pairwise & 42.74            & 42.74          & 61.57            & 61.71 \\ \hline\hline\hline
			PosSeq+Pairwise+fc & \textbf{47.43}            & \textbf{47.43}          & \textbf{64.17}            & \textbf{64.86} \\  \hline 
			VTransE~\cite{zhang2016vtranse} & 44.76            & 44.76          &62.63    & 62.87 \\ \hline
		\end{tabular}}
	\end{table}

\subsection{Evaluations of Phrase \& Relation Detection}
\textbf{Comparing Methods}. We evaluate the overall performance of PPR-FCN for WSVRD. We compared the following methods: 1) \textbf{GroundR}~\cite{rohrbach2016grounding}, a weakly supervised visual phrase grounding method. We used the image-level triplets as the input short phrases for their language embedding LSTM; 2) \textbf{VisualPhrase-WSDNN}, its idea was originally proposed in~\cite{sadeghi2011recognition} that considers a whole relation triplet as a class label. As it can be reduced to a weakly supervised object detection task, we used WSDDN~\cite{bilen2016weakly} pipeline to implement VisualPhrase. Note that our parallel FCN architecture cannot be adopted in VisualPhrase since the number of relation classes is too large to construct the conv-filters. 3) \textbf{VTransE-MIL}, we followed the same pipeline of VTransE~\cite{zhang2016vtranse} but using the NoisyOR Multiple Instance Learning (MIL)~\cite{maron1998framework} as the loss function for object and relation detections. 4) \textbf{PPR-FCN-single}, we only use the classification branch to implement PPR-FCN. We also compared three fully supervised models as baselines VTransE\cite{zhang2016vtranse}, Lu's-VLK~\cite{lu2016visual}, and Supervised-PPR-FCN, which is our proposed PPR-FCN applied in the supervised setting. Note that GroundR and VisualPhrase are based on joint boxes prediction and thus they can only perform phrase detection.

\begin{figure}
	\centering
	\includegraphics[width=.7\linewidth]{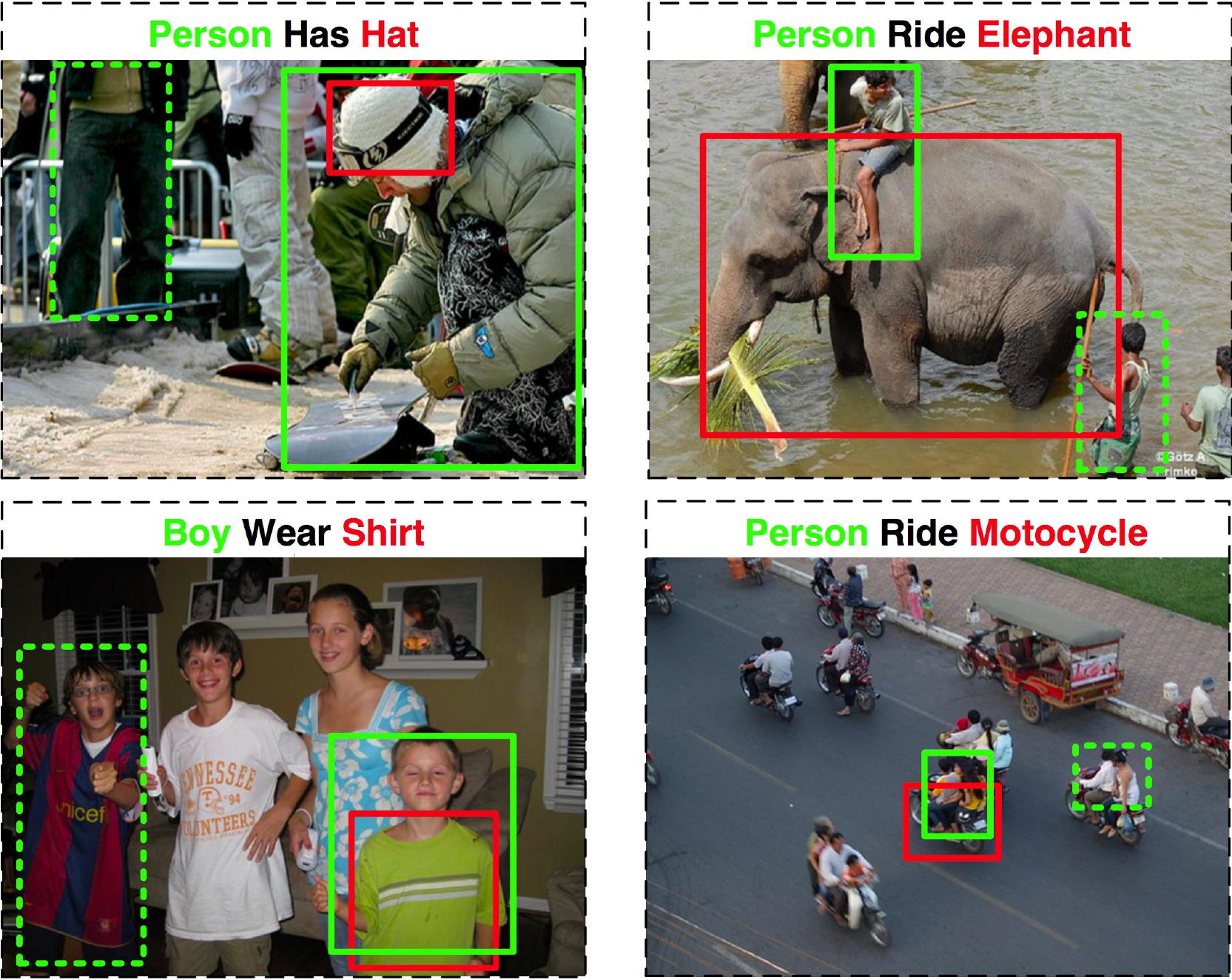}
	\caption{Qualitative examples of relation detection on VG. Compared to the results of PPR-FCN (solid green bounding boxes), VTransE-MIL (dashed green bounding boxes) is more likely to misalign \texttt{subject} to \texttt{object} if there are multiple instances of \texttt{subject}.}
	\label{fig:6}
\end{figure}

\textbf{Results}
Table~\ref{tab:3} and~\ref{tab:4} reports the phrase and relation detection performances of various methods. We have the following observations:
\\ 
1) For phrase detection, GroundR and VisualPhrase-WSDDN perform much poorly than VTransE-MIL and PPR-FCN. The reason is two-fold. First, EdgeBox is not designed to generate joint proposals of two interacted objects and thus we limited the number of proposals to 300 to handle 90,000 region pairs, where the top 300 proposals may be low recall of objects. Second, as discovered in~\cite{lu2016visual,zhang2016vtranse}, once we consider the relation as a whole class label, the training samples for each relation class are very sparse, which will worsen the training of WSPP.
\\
2) PPR-FCN outperforms VTransE-MIL in both phrase and relation detections. The reason is that VTransE does not explicitly model the spatial context in relation modeling while our PPR-FCN does. Note that this is crucial since the context can remove some incorrect \texttt{subject}-\texttt{object} configurations, especially when the supervision is only at the image level. For example, Figure~\ref{fig:6} shows that the position-role-sensitive score map and pooling design in PPR-FCN can correct misaligned \texttt{subject} and \texttt{object} when there are multiple instances.
\\ 
3) Our parallel design of PPR-FCN is significantly better than its counterpart PPR-FCN-single. This demonstrates that for weakly supervised learning with many candidate instances (\ie, region pairs), the parallel design without parameter sharing can prevent from bad solutions.
\\
4) There is a large gap between WSVRD and supervised VRD, \eg, PPR-FCN can only achieve  less than a half of the performance of supervised VRD such as Supervised-PPR-FCN and VTransE. We believe that the bottleneck is mainly due to the WSOD module that tends to detect small discriminative part instead of the whole object region. As shown in Figure~\ref{fig:5}, most of the failed relation detection is due to the failure of object detection. For example, for large and background-like objects such as \texttt{mountain}, \texttt{sky} and \texttt{building}, only small regions are detected; for \texttt{tower}, only the most discriminative ``spire'' is detected.
\\
5) Even though the fully-connected subnetworks is very helpful in predicate prediction as we discussed in Section~\ref{sec:4.4}, Supervised-PPR-FCN can still outperform the fc-based VTransE due to the effectiveness of the pairwise RoI pooling, which can correct wrong spatial context (Figure~\ref{fig:6}) Note that since PPR-FCN is designed for WSVRD, we cannot remove bad RoI pairs using pairwise groundtruth bounding boxes, which may lead to significant improvement in supervised settings~\cite{li2017vip}.
\\
6) Thanks to the FCN architecture introduced in PPR-FCN, it can not only speed up the WSOD, but also efficiently handle tens of thousands region pairs in WSVRD. For example, as reported in Table~\ref{tab:5}, PPR-FCN is about 2$\times$ faster than VTransE-MIL using per-region fc subnetworks. It is worth noting that the number of parameters of PPR-FCN is much smaller that VTransE-MIL (\eg, millions of fc parameters) as we only have $\mathcal{O}(k^2(C+1+R))$ conv-filters. Our current bottleneck is mainly due to the EdgeBox~\cite{zitnick2014edge} proposal generation time, as we strictly stick to the weak supervision setting that any module should not exploit bounding boxes. However, in practice, we can use generic class-agnostic RPN~\cite{ren2015faster} to generate proposals in 100 ms/img. 
\begin{table}[t]
	\centering
	\caption{Phrase detection performances (R@K\%) of various methods in weakly supervised and supervised settings (bottom three rows) on VRD and VG. }
	\label{tab:3}
	\scalebox{.8}{
		\begin{tabular}{|c|c|c||c|c|}
			\hline
			Dataset     & \multicolumn{2}{c||}{VRD} & \multicolumn{2}{c|}{VG} \\ \hline
			Metric    & R@50              & R@100            & R@50            & R@100            \\ \hline
			GroundR~\cite{rohrbach2016grounding}  & 0.15            & 0.18             & 0.33           & 0.81 \\ \hline
			VisualPhrase-WSDDN & 0.26            & 0.37          & 0.21            & 0.78 \\ \hline
			VTransE-MIL & 4.09            & 6.15          & 1.53           & 2.02 \\ \hline
			PPR-FCN-single & 3.56 & 4.31 & 0.87 &0.98 \\ \hline
			PPR-FCN &\textbf{6.93} 	& \textbf{8.22}	&\textbf{2.41} 	&\textbf{3.23}\\\hline\hline\hline
			Lu's-VLK~\cite{lu2016visual} & 16.17            & 17.03          & --         & -- \\ \hline
			VTransE~\cite{zhang2016vtranse} & 19.42            & 22.42          & 9.46    & 10.45 \\ \hline
			Supervised-PPR-FCN & \textbf{19.62}            & \textbf{23.15}          & \textbf{10.62}    & \textbf{11.08} \\ \hline
		\end{tabular}}
		\vspace{-6mm}
	\end{table}
	
\begin{table}[t]
	\centering
	\caption{Relation detection performances (R@K\%) of various methods in weakly supervised and supervised settings (bottom three rows) on VRD and VG.}
	\label{tab:4}
	\scalebox{.8}{
		\begin{tabular}{|c|c|c||c|c|}
			\hline
			Dataset     & \multicolumn{2}{c||}{VRD} & \multicolumn{2}{c|}{VG} \\ \hline
			Metric    & R@50              & R@100            & R@50            & R@100            \\ \hline
			VTransE-MIL & 4.28            & 4.54          & 0.71            & 0.90 \\ \hline
			PPR-FCN-single &3.56 &4.15 &1.08 &1.63 \\ \hline
			PPR-FCN &\textbf{5.68} 	&\textbf{6.29} 	&\textbf{1.52} 	&\textbf{1.90}\\\hline\hline\hline
			Lu's-VLK~\cite{lu2016visual} & 13.86            & 14.70          & --         & -- \\ \hline
			VTransE~\cite{zhang2016vtranse} & 14.07          & 15.20        & 5.52    & 6.04 \\ \hline
			Supervised-PPR-FCN & \textbf{14.41}           & \textbf{15.72}          & \textbf{6.02}    & \textbf{6.91} \\ \hline
		\end{tabular}}
	\end{table}

\begin{table}[t]
	\centering
	\caption{Titan X GPU test time (ms/img) of the fc subnetwork based weakly supervised method, VTransE-MIL and PPR-FCN (excluding the proposal generation time cost by EdgeBox, which is 700 ms/img). Both VTransE-MIL and PPR-FCN adopts ResNet-50 as the base CNN and 100 detected object proposals, \ie, 10,000 region pairs for predicate prediction.}
	\label{tab:5}
		\scalebox{.8}{
	\begin{tabular}{|c|c|}
		\hline
		 VTransE-MIL & PPR-FCN\\ \hline
		  270          & 150     \\ \hline
	\end{tabular}
}
\vspace{-4mm}
\end{table}

\section{Conclusion}
We presented a parallel, pairwise region-based, fully convolutional network: PPR-FCN, for the challenging task of weakly supervised visual relation detection (WSVRD). PPR-FCN has two novel designs towards the optimization and computation difficulties in WSVRD: 1) PPR-FCN is a parallel FCN network for simultaneous classification and selection of objects and their pairwise relations, and 2) the position-role-sensitive conv-filters and pairwise RoI pooling that captures the spatial context of relations. Thanks to the shared computation on the entire image, PPR-FCN can be efficiently trained with a huge amount of pairwise regions.  PPR-FCN provides the first baseline for the novel and challenging WSVRD task, which can foster practical visual relation detection methods for connecting computer vision and natural language. We found that the bottleneck of PPR-FCN is the WSOD performance. Therefore, future research direction may focus on jointly modeling WSOD and WSVRD by incorporating relations as the contextual regularization for objects.

{\footnotesize
\bibliographystyle{ieee}
\bibliography{egbib}
}

\end{document}